\documentclass[10pt,twocolumn,letterpaper]{article}

\usepackage[pagenumbers]{cvpr} 

\usepackage{graphicx}
\usepackage{amsmath}
\usepackage{amssymb}
\usepackage{booktabs}
\usepackage{multirow}

\usepackage[pagebackref,breaklinks,colorlinks]{hyperref}

\usepackage[capitalize]{cleveref}
\crefname{section}{Sec.}{Secs.}
\Crefname{section}{Section}{Sections}
\Crefname{table}{Table}{Tables}
\crefname{table}{Tab.}{Tabs.}

\def\cvprPaperID{3023} 
\def\confName{CVPR}
\def\confYear{2022}

\begin{document}
\title{3D Adversarial Attacks Beyond Point Cloud}

\author{\textbf{Jinlai Zhang\textsuperscript{\rm 1\thanks{Equal Contribution}}, Lyujie Chen\textsuperscript{\rm 2\footnotemark[1]}, Binbin Liu\textsuperscript{\rm 2}, Bo Ouyang\textsuperscript{\rm 2}, Qizhi Xie\textsuperscript{\rm 2}}\\
\textbf{Jihong Zhu\textsuperscript{\rm 1,2}, Weiming Li\textsuperscript{\rm 3}, Yanmei Meng\textsuperscript{\rm 1}}\\
\textsuperscript{\rm 1}Guangxi University~~
\textsuperscript{\rm 2}Tsinghua University~~
\textsuperscript{\rm 3}Samsung Research China - Beijing (SRC-B)~~\\
\small \{cuge1995\}@gmail.com, \{clj19,lbb19,oyb19,xqz20\}@mails.tsinghua.edu.cn
\\
\small \{jhzhu\}@mail.tsinghua.edu.cn \{weiming.li\}@samsung.com
\{gxu\_mengyun\}@163.com}

\maketitle
\begin{abstract}
Recently, 3D deep learning models have been shown to be susceptible to adversarial attacks like their 2D counterparts. Most of the state-of-the-art (SOTA) 3D adversarial attacks perform perturbation to 3D point clouds. To reproduce these attacks in the physical scenario, a generated adversarial 3D point cloud need to be reconstructed to mesh, which leads to a significant drop in its adversarial effect. In this paper, we propose a strong 3D adversarial attack named Mesh Attack to address this problem by directly performing perturbation on mesh of a 3D object. In order to take advantage of the most effective gradient-based attack, a differentiable sample module that back-propagate the gradient of point cloud to mesh is introduced. To further ensure the adversarial mesh examples without outlier and 3D printable, three mesh losses are adopted.
Extensive experiments demonstrate that the proposed scheme outperforms SOTA 3D attacks by a significant margin.
We also achieved SOTA performance under various defenses. 
Our code will available online.

\end{abstract}
\section{Introduction}
\label{sec:intro}
Three-dimensional (3D) shape representation is one of the most fundamental topics in computer vision. As an expressive 3D representation, \textit{point clouds} can be directly output from 3D sensors like LiDAR or Kinect, thus becoming the primary data structure in various 3D computer vision tasks. \textit{Mesh} is another practical 3D representation that is widely used in the computer graphics and computer-aided design (CAD) field but receives less attention in the computer vision field. With the help of 3D printing technology, mesh has become a bridge connecting digital space and physical space~\cite{knnattack,geoa3}.

\begin{figure}[!thp]
    \centering
    \includegraphics[width=\linewidth]{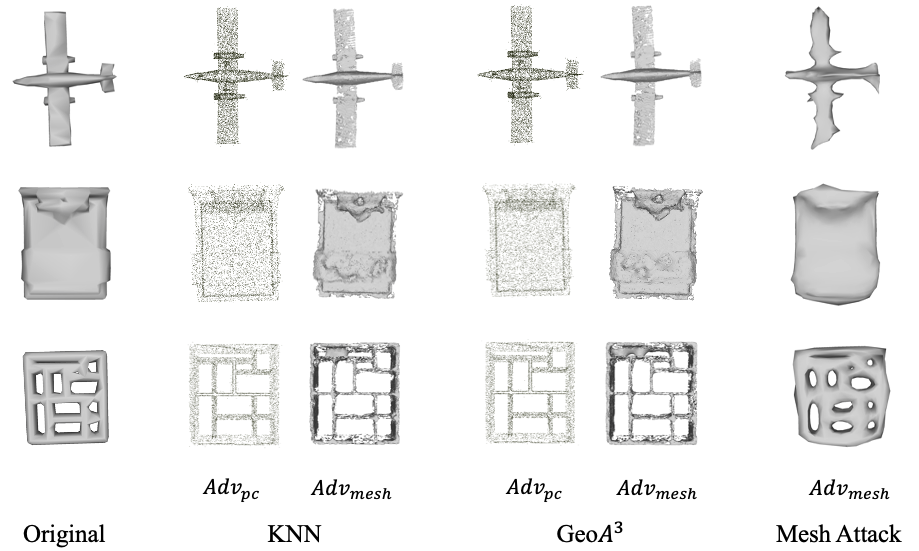}
    \caption{
    \textbf{Motivation.} Adversarial examples crafted by the KNN attack~\cite{knnattack}, the $GeoA^3$ attack~\cite{geoa3} and our proposed Mesh Attack on the Pointnet~\cite{pointnet}. The $Adv_{pc}$ and $Adv_{mesh}$ denote the adversarial point cloud and the adversarial mesh, respectively. The $Adv_{mesh}$ of KNN attack and $GeoA^3$ attack were reconstructed from the $Adv_{pc}$. The adversarial perturbation
was small in the point cloud space, but lead to very substantial “noise” in the reconstructed mesh. Our Mesh Attack is able to avoid those "noise".
    }
    \label{compare}
    \setlength{\belowcaptionskip}{-1cm}
\end{figure}

\begin{figure*}
    \centering
    \includegraphics[width=.75\linewidth]{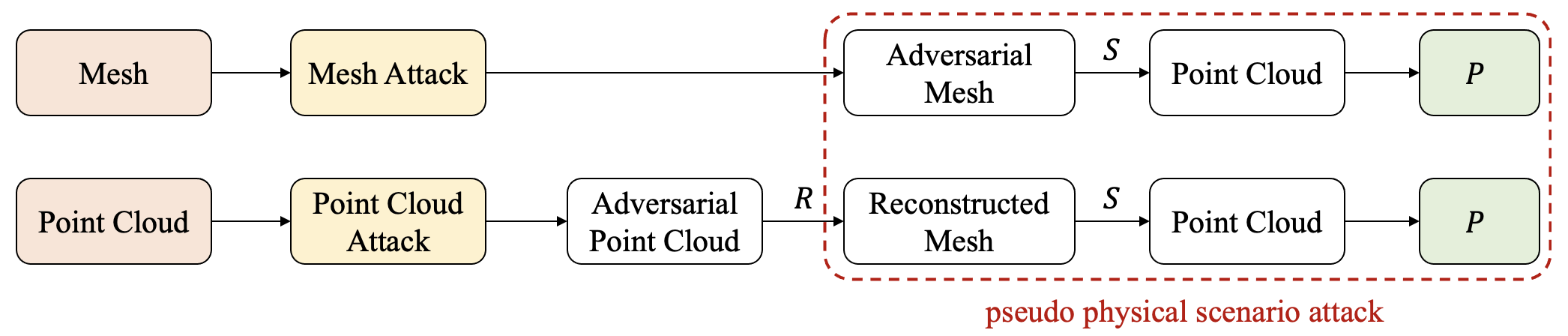}
    \caption{
    \textbf{Comparison between our Mesh Attack (top row) and previous point cloud attacks (bottom row) in pseudo physical scenarios.} $R$ denotes the process of reconstructing mesh from the point cloud. $S$ denotes the process of the sampling point cloud from mesh surface. $P$ denotes point cloud classifiers. The adversarial effect of previous point cloud attacks degrades significantly, while ours remains.
    }
    \label{main_diff}
    \setlength{\belowcaptionskip}{-1cm}
\end{figure*}

To better understand the 3D objects, a series of works~\cite{pointnet,pointconv,pointnet,pointnet++,dgcnn,rscnn} that processed point clouds through deep neural networks (DNNs) have been proposed, which further inspired the works in downstream tasks such as 3D object detection~\cite{pvrcnn}. With the development of DNNs, the threat of adversarial examples that fool the models has also gained increasing attention~\cite{adreview,adsurvey}. Adversarial examples are imperceptible to humans but can easily fool DNNs in the testing stage~\cite{yuan2019adversarial}, which have been extensively studied in the 2D image field~\cite{fgm,mifgm,pgd,ifgm,advnlp,zooattackblack,tu2019autozoomblack}. Although a series of works were proposed, 3D adversarial attacks are still relatively under-explored and suffer several problems. 
The major problem is that the existing works are barely physically realizable. They mostly perform adversarial attacks on point clouds data by adding or deleting some points or changing their coordinates~\cite{knnattack,geoa3,generatingadpoint,pointcloudsaliencymaps,lggan,advpc,attsordefense}. Such attacks are limited to point cloud space in the digital world and are difficult to apply to the real environment, where point clouds are either generated by 3D scanners or sampled from reconstructed meshes.
To perform real-world adversarial attacks, the KNN attack~\cite{knnattack} and the $GeoA^3$ attack~\cite{geoa3} made some early trials. They reconstructed the mesh of the generated adversarial examples by a point-to-mesh reconstruction algorithm and then obtained the point cloud by direct sampling from the mesh surface. We call this kind of attack as \textit{pseudo physical scenario attack}. 
The results~\cite{knnattack,geoa3} showed that the adversarial effect of the $GeoA^3$ attack surpassed the KNN attack, but the performance of adversarial point cloud examples sampled from the reconstructed mesh degraded significantly compared to the input point cloud. The attack success rate dropped from 100\%  to under 10\% in most cases. 
Furthermore, they also performed \textit{real physical scenario attack} by sampling point clouds from the 3D-printed objects with a 3D scanner, those objects are randomly selected from the adversarial mesh that successfully fool the point cloud classifier in the pseudo physical scenario to ensure the attack success rate in the real physical scenario. In this situation, the attack performance further declined. This demonstrated that the real-world point cloud attack still remains a challenging task~\cite{geoa3}. 
To figure out the reason for the degradation of attack performance, we present the perturbed point cloud and its counterpart mesh of the KNN attack~\cite{knnattack} and the $GeoA^3$ attack~\cite{geoa3} in Figure~\ref{compare}.
The visualization suggests that adversarial perturbations, while small in the point cloud space, lead to very substantial “noise” in the reconstructed mesh. This not only caused a significant loss of adversarial information but also made it very difficult to 3D print the reconstructed mesh.


To this end, in this paper, we propose Mesh Attack, a 3D adversarial attack that performs perturbation on the mesh data instead of point clouds. 
Figure~\ref{main_diff} shows the major difference between our Mesh Attack and previous point cloud attacks.  Since the attack is directly applied to the mesh, the "noise" in the reconstructed mesh in previous works~\cite{knnattack,geoa3} is avoided. To back-propagate the gradient of point cloud to mesh, a differential sample module is introduced. 
However, directly adding perturbations to the mesh will cause self-intersection and flying vertices problems, which introduce a lot of outliers. This not only makes fatal errors during the 3D printing but also be easily defended by Statistical Outlier Removal (SOR) method~\cite{dupnet} that removes outlier points with a large kNN distance. 
To further ensure the adversarial mesh examples without outlier and 3D printable, three mesh losses are adopted, \ie, the chamfer loss to regress the mesh vertices close to their correct position, and the laplacian loss and the edge length loss regularize the mesh to be smoother and avoid the mesh self-intersection and flying vertices problems. 
By confining the mesh perturbation to smooth shape transition, outlier points are eliminated. This not only makes our attack appear more natural to human beings but also more difficult to be defended by current point cloud defense algorithms. 
Due to the difficulty to compare SOTA 3D attacks on a large scale in \textit{real physical scenario attack}, we followed the previous settings~\cite{knnattack,geoa3} and compared SOTA 3D attacks in  \textit{pseudo physical scenario attack}.
Extensive experimental results show that our Mesh Attack outperforms SOTA 3D attacks by a significant margin, which makes a significant step toward real-world point cloud attack. To further validate the effectiveness of our Mesh Attack, we performed \textit{real physical scenario attack}, the results show that most of the adversarial meshes generated by Mesh Attack can attack successfully  after scanning. Also, the adversarial robustness of our proposed Mesh Attack outperforms the SOTA point cloud attack algorithms by a large margin under various defenses across different victim models in the pseudo physical scenario. Furthermore, since the adversarial mesh examples are directly generated by Mesh Attack, they enable a black-box attack to MeshNet~\cite{feng2019meshnet}, which achieves strong attack performance. To the best of our knowledge, Mesh Attack is the first adversarial attack on the mesh classifier. 
Our Mesh Attack revealed the robustness issues for mesh-related 3D data representation and analysis, which we hope to bring inspiration for future real-world 3D AI researches and applications.

The \textbf{main contributions} of this paper can be summarized as: (1) We present a strong 3D adversarial attack method named Mesh Attack that can directly generate adversarial mesh examples. From the mesh examples, adversarial point clouds can be sampled that fool point cloud classifiers. (2) Extensive experiments show that our Mesh Attack outperforms the existing point cloud attack methods by a large margin in the pseudo physical scenario and hard to be defended by the current state-of-the-art point cloud defense algorithms. (3) Mesh Attack is also shown to work as the world-first 3D adversarial attack on mesh classifier in the black-box setting.

%


\section{Related works}
\label{sec:related}


\paragraph{Deep learning in 3D}
With the popularity of 3D sensors and the needs of real-world autonomous driving cars and robotics, 3D deep learning has surged in recent years. 3D data has many representations, \ie, multi-view RGB(D) images, volumetric, mesh, point clouds, and primitive-based CAD models. Although multi-view images and volumetric are regular grids~\cite{su2015multi,qi2016volumetric_multi} that can directly apply convolution neural networks (CNN), the high computational cost limits their applications. Point clouds represent the real world as a set of points. After the breakthrough made by Pointnet~\cite{pointnet}, point clouds have become popular in the 3D computer vision community~\cite{pointasnl,pointnet++,dgcnn,rscnn,pointconv,pointcutmix}. However, there is no connection between each point, thus lacking the surface information of the object. In comparison, the mesh representation has the characteristics of lightweight and rich shape details. Therefore, enabling DNNs to process mesh would be promising. MeshNet~\cite{feng2019meshnet} firstly proposed mesh convolution to process mesh data, which performs well in 3D shape classification and retrieval. Recently, Joseph~\etal~also introduced BRepNet~\cite{lambourne2021brepnet} to segment primitive-based CAD models.

\paragraph{Adversarial attacks in 3D}
DNNs are vulnerable to adversarial examples, which have been extensively explored in the 2D image domain~\cite{firstadv,moosavi2016deepfool,moosavi2017universal,su2019one}. Recently, the adversarial attack has also gained attention from the 3D vision community. Among existing representations of 3D data, only the adversarial robustness of point clouds has been widely studied. \cite{generatingadpoint} first proposed to generate adversarial point clouds examples. However, the generated adversarial point clouds examples are very messy and have point outliers, which can be easily perceivable by humans. To address the issue of point outliers, the kNN attack~\cite{knnattack} adopted a kNN distance constraint, a clipping, and a projection operation to generate a more smooth and imperceptible adversarial point cloud example. The Geometry-Aware Adversarial Attack ($GeoA^3$) ~\cite{geoa3} further improved the imperceptible to humans. The perturbation-based attacks can be removed by the statistical outlier removal (SOR)~\cite{dupnet} method that removes points with a large kNN distance if the perturbation is too large. To solve this problem, \cite{attsordefense} developed the JGBA attack, which is an efficient attack to SOR defense. Besides, the point drop attack~\cite{pointcloudsaliencymaps} was developed by a gradient-based saliency map, which iteratively removes the most important points. AdvPC~\cite{advpc} improved the transferability of adversarial point cloud examples by utilizing a point cloud auto-encoder. LG-GAN~\cite{lggan} utilized the powerful GANs~\cite{goodfellowgan} to generate adversarial examples guided by the input target labels. 
Although the existing attacks have a high attack success rate but cannot guaranty that the generated adversarial examples are physically realizable. In this paper, we propose Mesh Attack to solve this problem.

\paragraph{Adversarial attacks in physical world}
A series of research has been conducted to generate adversarial examples in the physical world. Kurakin~\etal~\cite{adv_inphy} firstly shown that adversarial examples can still be effective via printing and recapturing using a cell phone camera. Athalye~\etal~\cite{athalye2018synthesizingphy} further improved the robustness of physical adversarial attack under natural noise, distortion, and affine transformation. The AdvPatch attack~\cite{brown2017adversarial_patch_phy} 
fooled DNNs by pasting an adversarial patch on the printed picture. Inspired by this work, the adversarial eye-glass ~\cite{sharif2016accessorizephy} and adversarial stickers in traffic sign~\cite{robustphy} were proposed. Moreover, the adversarial t-shirts~\cite{tshirt_phy} made human stealth in the object detector and the Adversarial Camouflage~\cite{camouflagephy} generated natural style adversarial examples in the physical world. 
However, the attack algorithms mentioned above focus on fooling a 2D image classifier or 2D image object detector, limiting their application in the 3D area. To generate 3D adversarial examples, Tu~\etal~\cite{tu2020physically3d} added the adversarial mesh on the car's roof, which can fool the 3D point cloud object detector. But they concentrate on the 3D point cloud object detector, we focus on the 3D point cloud classifier.
Another way to perform the 3D attack in the physical world is LiDAR spoofing attacks~\cite{cao2019adversarialcars3d}, but this is hard to implement in real driving conditions due to the LiDAR spoofer need to aim at the LiDAR.

\section{Method}

\subsection{Problem Overview}

\begin{figure*}
    \centering
    \includegraphics[width=0.8\linewidth]{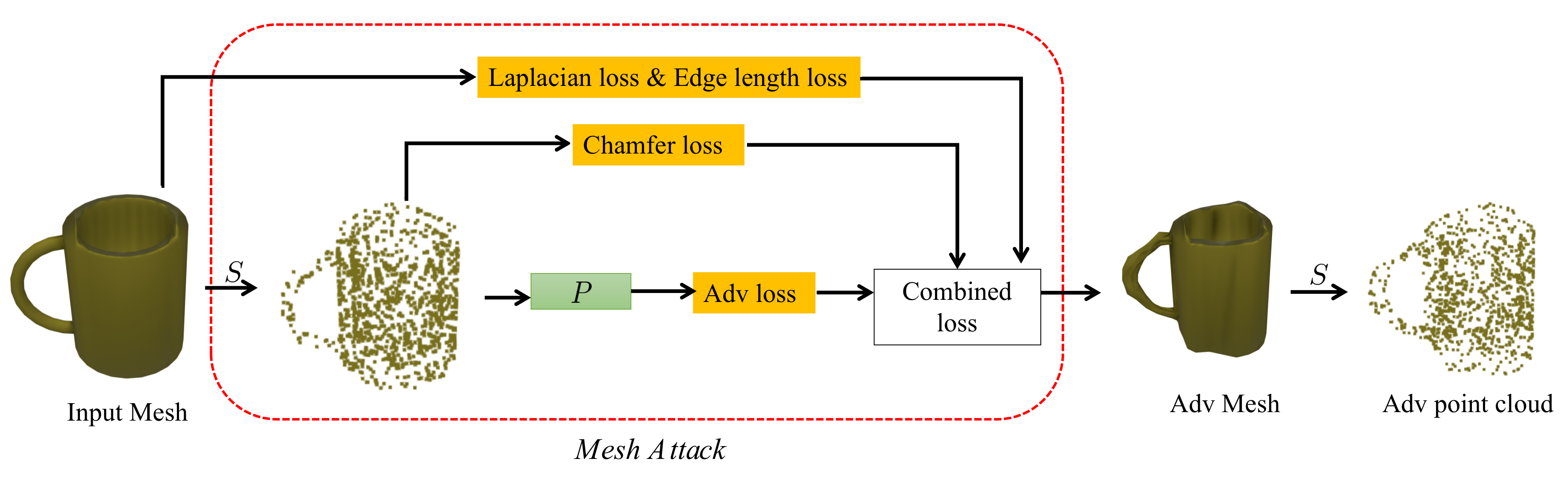}
    \caption{\textbf{Our attack pipeline.} $S$ denotes the differential sample module, $P$ denotes point cloud classifiers, Adv loss denotes the adversarial loss, Adv Mesh denotes the adversarial mesh examples, Adv point cloud denotes the adversarial point cloud examples.}
    \label{pipeline}
\end{figure*}

The pipeline of Mesh Attack is illustrated in Figure~\ref{pipeline}. It consists of a differential sample module $S$ and a point cloud classifier $P$. Let $M$ denote the original mesh, represent as $M=(V, F)$, where $V \in \mathbb{R}^{N_{V} \times 3}$ is the $xyz$ coordinates of vertices and $F \in \mathbb{Z}^{N_{F} \times 3}$ is the set of triangle faces and encodes each triangle with the indices of vertices. $N_{V}$ and $N_{F}$ denote the number of vertices and triangles, respectively.
$x$ denotes its sampled point cloud and $y$ denotes the corresponding true label. $S$ is to sample point cloud $x$ from a given mesh $M$ in a differentiable way.
Our Mesh Attack keeps the triangle topology of the original mesh fixed and only updates the vertex positions. Formally, for a point cloud classifier $P(x):x \rightarrow y$, the objective of our attack method is to find a perturbation $\Delta$ and generate an imperceptible adversarial mesh example $M^{adv}=\left(V^{adv}, F\right)$, where $V^{adv}=V+\Delta$, and further samples the adversarial point cloud example $x^{adv}$ to fool the classifier $P$.

\subsection{Mesh Attack}
As shown in Figure~\ref{pipeline}, different from previous point attack algorithms, our adversarial point cloud examples are sampled from the adversarial mesh examples and $\Delta$ is optimized by minimizing the loss. 

\textbf{Differential  sample  module.} In order to optimize the $\Delta$ via back-propagation directly, the sampling process of sample points from mesh need to be differentiable. However, making sampling points from mesh differentiable is hard. To tackle this challenge, we follow the rule of Triangle Barycentric Coordinates. Since the mesh we used is triangle mesh, we can sampling points from mesh's triangle faces in a differentiable way.

For a given triangle $P_{ABC}$, its vertices are $v_1$, $v_2$ and $v_3$, respectively. For any point $p_t$ inside $P_{ABC}$, the barycentric coordinates are given as follows:
\begin{equation}
P_t = \lambda_1*v_1 + \lambda_2*v_2 + \lambda_3*v_3
\end{equation}
where $\lambda_1+\lambda_2+\lambda_3=1$, in this way, we can sampling points from mesh in a differentiable way.

\textbf{Mesh Attack Losses.} To optimize the $\Delta$, we follow the C\&W attack framework to minimize the loss. Our Mesh Attack's loss can be divided into two parts, one is a cross-entropy loss that promotes the misclassification of adversarial point cloud examples, another one is mesh losses that constrained the adversarial mesh examples to be smooth. The cross-entropy loss is commonly used for classification tasks, it is defined as follows:
\begin{equation}
L_{M i s}=-\sum_{i}\sum_{c=1}^{m} y_{c} \log \left(p_{c}\right)
\end{equation}
where $m$ is the number of classes, $y_{c}$ is the ground truth label, $p_{c}$ is the predicted probability.
The mesh loss can be further depart as the chamfer loss, the laplacian loss and the edge length loss. The chamfer loss is commonly used in the mesh generation literature~\cite{wen2019pixel2mesh++,wang2018pixel2mesh,wang2020pixel2mesh}, it regress the vertices in a mesh to its correct position. It is defind as follows:
\begin{align}
\begin{split}
l_{\mathrm{c}}\left(S_{1}, S_{2}\right)&=\frac{1}{S_{1}} \sum_{x \in S_{1}} \min _{y \in S_{2}}\|x-y\|_{2}^{2}\\
&+\frac{1}{S_{2}} \sum_{y \in S_{2}} \min _{x \in S_{1}}\|x-y\|_{2}^{2}
\end{split}
\end{align}

where $S_{1}=S(M)$, $S_{2}=S(M^{adv})$ represent two sets of 3D point clouds, respectively. The first item represents the sum of the minimum distance between any point $x$ in $S_{1}$ and $S_{2}$. The second term means the sum of the minimum distance between any point $y$ in $S_{2}$ and $S_{1}$. In this paper, we use the chamfer loss to constrain the geometry distance between the original input mesh and the generated adversarial mesh examples. If this distance is larger, the difference between the meshes is greater; if the distance is smaller, the perceiveness effect is better.
However, only use the chamfer loss makes the generated adversarial mesh examples self-intersection, and sometimes flying vertices. To handle these problem, we further introduced the laplacian loss and the edge length loss. The laplacian loss can be defind as follows:
\begin{equation}
\delta_{i}=V_i-\sum_{k \in \mathcal{N}(i)} \frac{1}{\|\mathcal{N}(i)\|} V_k
\end{equation}
\begin{equation}
l_{lap}=\sum_{i}\left\|\delta_{i}\right\|_{2}^{2}
\end{equation}
where each vertex $V_i \in M^{adv}$ is denoted as $ V_{i}=\left(x_{i}, y_{i}, z_{i}\right) $, and $\delta_{i}$ is the laplacian coordinate for $V_i$ that compute the difference between $V_i$ and the center of mass of its immediate neighbors.The edge length can be defind as follows:
\begin{equation}
l_{edge}=\sum_{\theta_{i} \in e_{i}}\left(\cos \theta_{i}+1\right)^{2}
\end{equation}
where $\theta_{i}$ denotes the angle between the faces that
have the common edge $e_i$. Consequently, the regularizer term can be formulated as:
\begin{equation}
l_{reg}=\lambda_{1} l_{c}+\lambda_{2} l_{l a p}+\lambda_{3} l_{edge}
\end{equation}
where $\lambda_{1}$, $\lambda_{2}$ and $\lambda_{3}$ are the weighting parameters. 

In order to fair compare with the KNN attack~\cite{knnattack} and the $GeoA^3$ attack~\cite{geoa3}, we also follow the C\&W attack's optimization framework. So we first give a brief introduction of C\&W attack algorithm.
C\&W~\cite{cwattack} turns the process of obtaining adversarial examples into an optimization problem as
\begin{equation}
\min _{x^{adv}} L_{M i s}\left(x^{adv}\right)+c \cdot L_{R e g}\left(x^{adv}, x\right)
\end{equation}
where $L_{M i s}\left(x^{adv}\right)$ can promote the misclassification of $x^{adv}$, $L_{R e g}\left(x^{adv}, x\right)$ is a regularization term that minimize the distance between $x^{adv}$ and $x$, $c$ is automated adjusted by binary search. Thus, the optimization process of our Mesh Attack can be described as follows:
\begin{equation}
\min _{M^{adv}} L_{M i s}\left(S(M^{adv})\right)+c \cdot L_{R e g}\left(M^{adv}, M\right)
\end{equation}
where $c$ is a trade-off between perceiveness and adversarial strength.

\textbf{Mesh Attack Strength. }
We also utilize the infinity norm $
l_{\infty} $ to limit the strength of the perturbations. It constrained the perturbation $\left\|\Delta_{i}\right\|_{2}$ do not over a given threshold $l_{\infty}$, if the pertubation $\Delta_{i}$ is larger than the limit, it will update to $\Delta_{i}^{n e w}$ as follows:
\begin{equation}
\Delta_{i}^{n e w}=\frac{\Delta_{i}}{\left\|\Delta_{i}\right\|_{2}} \cdot l_{\infty}
\end{equation}

\section{Experiments}
In this section, we present the experimental results to demonstrate the effectiveness of the proposed method. We first specify the experimental settings in Sec. 4.1. In order to validate the design choice of Mesh Attack, we perform ablation studies in Sec. 4.2. Then we report the attack results of \textit{pseudo physical scenario attack} and \textit{real physical scenario attack}, and the robustness under defense in Sec. 4.3, Sec. 4.4 and Sec. 4.5, respectively. The black-box attack to the Mesh classifier is shown in Sec. 4.6. A Linux server with 8 Nvidia GTX 3090 GPUs is used to implement our experiments. 

\subsection{Setup}
We use the Manifold40~\cite{manifold40}, a variant of ModelNet40~\cite{modelnet40} dataset where all shapes are closed manifolds. We firstly select four commonly used networks in 3D computer vision community~\cite{pointcutmix,ifdefense} as the victim model, \ie, PointNet~\cite{pointnet}, PointNet++~\cite{pointnet++}, PointConv~\cite{pointconv}, and DGCNN~\cite{dgcnn}. We trained them from scratch, the input point cloud is randomly sampled from the normalized mesh surface during each epoch, and the test accuracy of each trained model is within 2\% of the best reported accuracy. We set $\lambda_{1}=1.0$, $\lambda_{2}=0.5$,  $\lambda_{3}=0.2$ and adversarial strength 0.1 in all experiments. We implement 10 steps binary search for $c$, the lower bound is 0 and the upper bound is 80, every step contains 1500 iterations to optimize the $\Delta$, and the Adam optimizer to minimize the loss. 
We implement the differentiable sampling module based on Pytorch~\cite{paszke2019pytorch} and  Pytorch3D~\cite{ravi2020pytorch3d}. We use meshes with 2000 faces and the sampling points is 1024 in the Sec. 4.3 and Sec. 4.4. The adversarial strength is 0.1.
In addition, in order to reduce the random sampling error, we conducted three independent experiments with different random seeds and randomly sampled five times on the generated adversarial mesh, and report the mean and variance of the experimental results in the table. We use attack success rate (ASR) as the evaluation metric.

\subsection{Ablation Study}
In this section, we conduct a series of experiments to help and validate the design choice of Mesh Attack. Pointnet is the victim model in this section. We follow \cite{generatingadpoint,geoa3} and randomly select $25$ instances for each of $10$ classes in the ModelNet40 testing set, which can be well classified by the Pointnet. The 10 classes are \textit{airplane, bed, bookshelf, bottle, chair, monitor, sofa, table, toilet,} and \textit{vase}. We use meshes with 8000 faces and the sampling points is 8000 in the ablation study.

\begin{figure*}[t]
  \centering
  \begin{minipage}{0.25\linewidth}
    \begin{subfigure}[t]{0.9\linewidth}
      \centering
      \includegraphics[width=\textwidth]{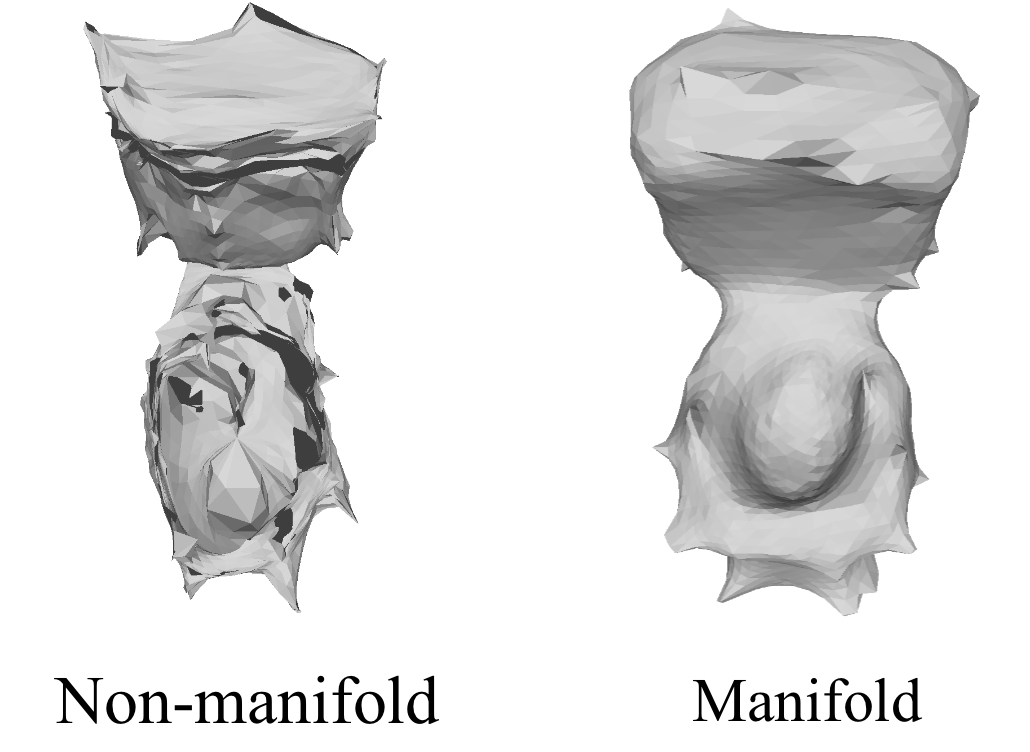}
      \caption{\textbf{Mesh Attack with manifold and non-manifold mesh.} The latter suffers the inconsistent face normal.}
      \label{manifold_or_not}
    \end{subfigure}
  \end{minipage}
  \begin{minipage}{0.36\linewidth}
    \begin{subfigure}[t]{0.9\linewidth}
      \includegraphics[width=\textwidth]{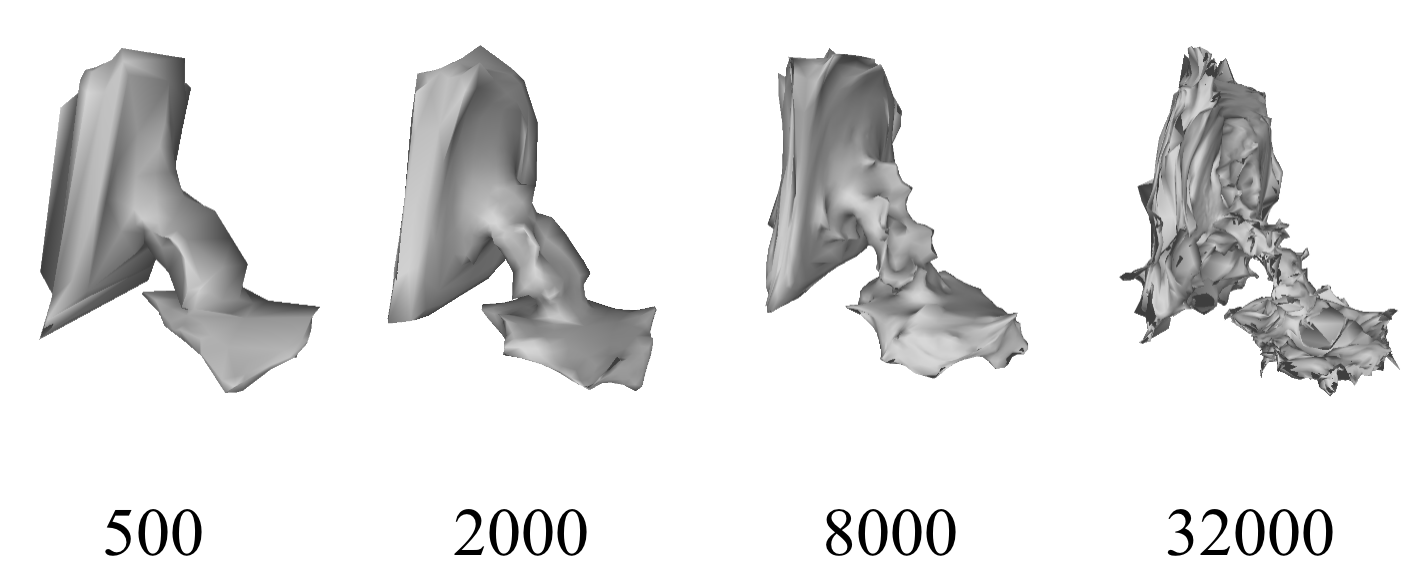}
      \caption{\textbf{Mesh Attack with different faces.} We perform the targeted attack with \textit{monitor} instance, the target is a \textit{curtain} and the faces are 500, 2000, 8000, and 32000 respectively.}
      \label{num_f}
    \end{subfigure}
  \end{minipage}
  \begin{minipage}{0.36\linewidth}
    \begin{subfigure}[t]{0.9\linewidth}
      \includegraphics[width=\textwidth]{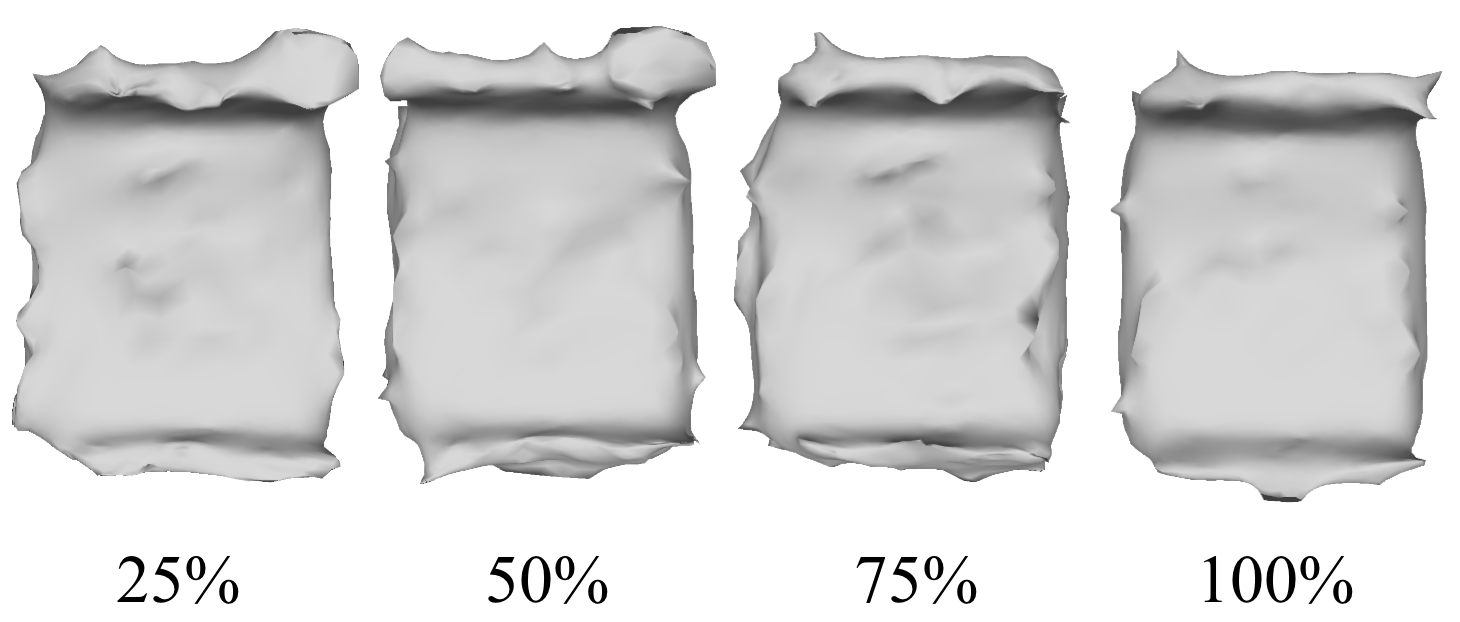}
      \caption{\textbf{Mesh Attack with different sample points.} We perform the targeted attack with \textit{bed} instance. The target is \textit{table} and the sample points are 2000, 4000, 6000 and 8000 respectively.}
      \label{num_points}
    \end{subfigure}
  \end{minipage}
  \begin{minipage}{0.48\linewidth}
    \begin{subfigure}[t]{0.9\linewidth}
      \includegraphics[width=\textwidth]{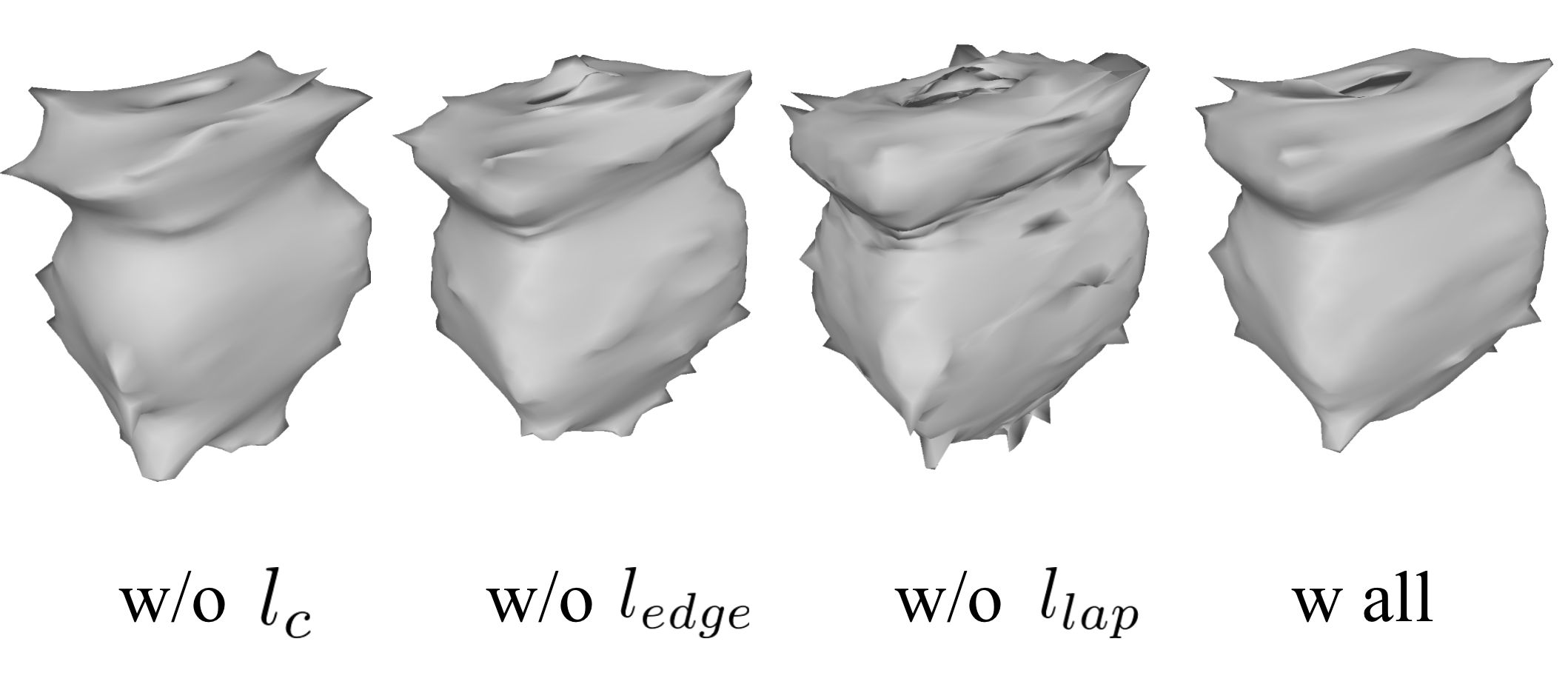}
      \caption{\textbf{Ablation of the three mesh losses.} We perform the targeted attack with \textit{vase} instance and the target is \textit{stairs}.}
      \label{meshlossab}
    \end{subfigure}
  \end{minipage}
  \begin{minipage}{0.48\linewidth}
    \begin{subfigure}[t]{0.9\linewidth}
      \includegraphics[width=\textwidth]{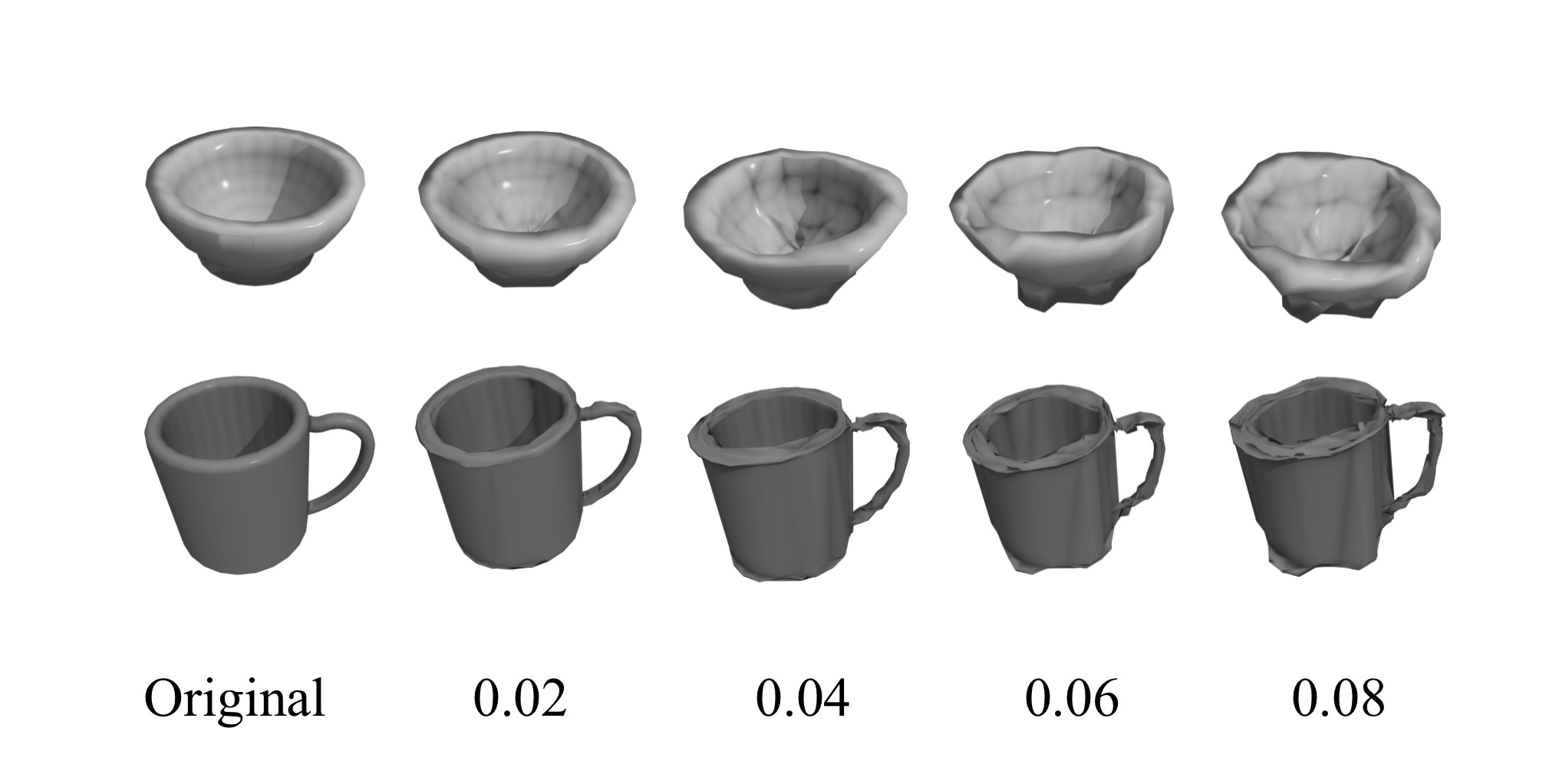}
      \caption{\textbf{Ablation of adversarial strength.}}
      \label{abstrenth}
    \end{subfigure}
  \end{minipage}
  \caption{\textbf{Visualization of ablation study.}}
  \label{aaa}
\end{figure*}

\textbf{The manifold mesh or not.} Manifold meshes are better for downstream tasks~\cite{manifold40}. However, in the ModelNet40, a lot of instances are not watertight or are non-manifold. We therefore investigate the effect of manifold mesh and non-manifold mesh on Mesh Attack. The ASR of manifold mesh and non-manifold mesh are 85.4\% and 57.2\%, respectively. The results in Figure \ref{manifold_or_not} indicate that non-manifold mesh is unsuitable for Mesh Attack. We therefore choose the manifold mesh in the following experiments.

\textbf{Number of faces per mesh.} The amount of detail of a mesh that represents a 3D object is determined by the number of faces on the mesh. We use meshes with 500, 2000, 8000, 32000 faces to implement our experiment. The ASR of four cases are 86.6\%, 96.3\%, 85.4\% and 46.7\%, respectively. The visualization of Mesh Attack with different faces are shown in Figure \ref{num_f}. The mesh with 32000 faces will cause unexpected shape deformation while using Mesh Attack, and mesh with faces less than 1000 might not represents complex object in rich details. We therefore recommend 2000-8000 faces for better ASR and perceptual quality when using Mesh Attack.


\textbf{Number of sampling points per mesh.} Since our differential sample module can only back-propagate the gradient of one point per face, we also study the effect of number of sampling points on Mesh Attack. We sampled 25\%, 50\%, 75\% and 100\% points of the total faces of a mesh during every iteration. The ASR of four cases are 97.2\%, 94.4\%, 91.7\% and 85.4\%, respectively. Figure \ref{num_points} shows the adversarial mesh generated with different sample points, we can see that when the sample points increased, the ASR decreased. However, the perceptual quality is increased with sample points.


\textbf{The impact of different loss.} Figure~\ref{meshlossab} illustrates four groups of adversarial examples without the three optional mesh losses. We can clearly see that the generated adversarial mesh examples with all mesh losses are smoother than those with partly mesh losses. The ASR of Mesh Attack without the chamfer loss,  the edge length loss and the laplacian loss are 96.6\%, 85.8\% and 74.7\%, respectively. 


\textbf{The impact of different adversarial strength.} Through the visual analysis of the adversarial strength in Figure \ref{abstrenth}, we can see that as the attack intensity increases, the deformation of the mesh increases. The ASR of the adversarial strength of 0.02, 0.04, 0.06 and 0.08 are 5.5\%, 23.8\%, 54.4\% and 74.1\%, respectively. When the adversarial strength is very small, it is very similar to the overused objects in our daily lives, and the surface has slight abrasion and deformation. 
Increasing the adversarial strength can increase the attack success rate, but flying vertices exist in most adversarial samples if the adversarial strength is too large in our Mesh Attack. Therefore, we do not recommend enormous adversarial strength.


\begin{table}[t]
\centering
\setlength{\tabcolsep}{3pt}
\renewcommand\arraystretch{1.15}
\caption{Evaluation of object-level adversarial effect via the attack success rates (\%) of randomly targeted attacks of kNN attack~\cite{knnattack}, $GeoA^3$ attack~\cite{geoa3} and Mesh Attack.}
\label{object_attack}
\vspace{2mm}
\resizebox{0.8\linewidth}{!}{
\begin{tabular}{c c c}
\toprule
Model & Method & Attack success rates (\%)\\
\midrule
\multirow{3}{*}{Pointnet} &
KNN Attack & 14.78
\\
& $GeoA^3$ Attack & 19.65
\\
&Mesh Attack &\bf90.23$\pm0.12$ 
\\
\midrule
\multirow{3}*{Pointnet++} &
KNN Attack &6.24
\\

&$GeoA^3$ Attack &11.20 
\\
&Mesh Attack &\bf98.03$\pm0.03$
\\
\midrule
\multirow{3}*{DGCNN} &
KNN Attack &4.17 
\\
&$GeoA^3$ Attack &8.24 
\\
&Mesh Attack &\bf59.44$\pm0.37$
\\
\bottomrule
\end{tabular}
}
\end{table}

\subsection{Pseudo physical scenario attack}
\begin{figure}[ht!]
    \centering
    \includegraphics[width=\linewidth]{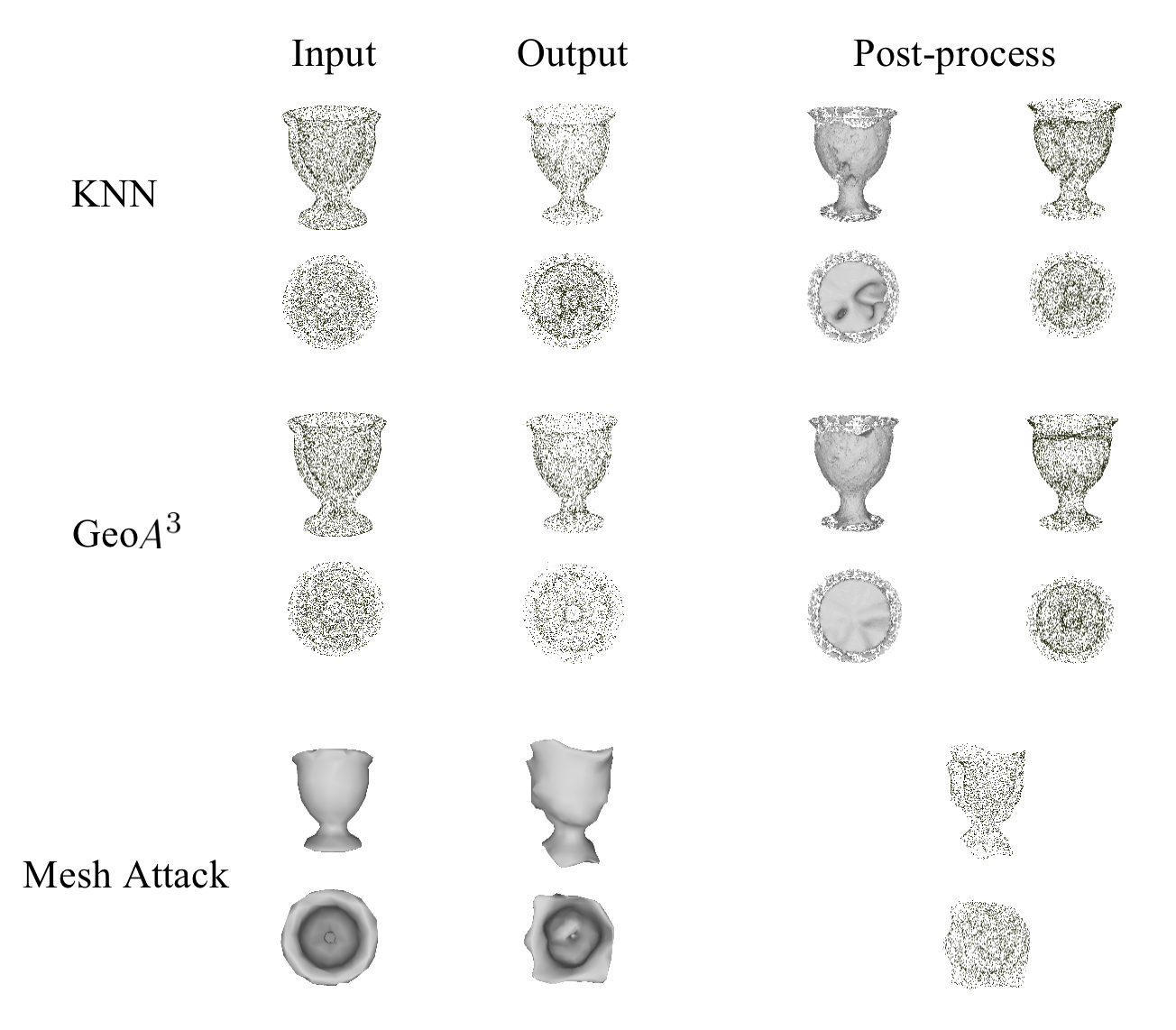}
    \caption{\textbf{Qualitative results on "vase" instance.} We provide the visual results from two perspectives, front view (top row) and top view (bottom row). For the KNN attack~\cite{knnattack} and $GeoA^3$ attack~\cite{geoa3}, the perturbation was small in the output adversarial point cloud, but lead to huge "noise" in the reconstructed mesh, and further affected the final adversarial point cloud in pseudo physical scenario. While for our Mesh Attack, the final adversarial point cloud can be directly sampled from the output adversarial mesh without quality drop.}
    \label{final_compare}
\end{figure}

Due to the difficulty to compare SOTA 3D attacks on a large scale in \textit{real physical scenario attack}, in this section, we followed the previous settings~\cite{knnattack,geoa3} and compared SOTA 3D attacks in  \textit{pseudo physical scenario attack}.
 Following the experiment detail in \cite{geoa3,knnattack}, we perform a randomly targeted attack of Mesh Attack and compare the object level adversarial effect of the KNN attack~\cite{knnattack} and $GeoA^3$ attack~\cite{geoa3} in the pseudo physical scenario where previous point cloud attacks~\cite{geoa3,knnattack} showing that attacks in this scenario are a challenging task, the results are shown in Table~\ref{object_attack}. The results of KNN attack~\cite{knnattack} and $GeoA^3$ attack~\cite{geoa3} are refer to ~\cite{geoa3}. 
Our Mesh Attack consistently outperforms the KNN attack and $GeoA^3$ attack by a significant margin in this scenario. 
This is a very impressive result because Mesh Attack was shown to be effective in this challenging task. 
Note that our Mesh Attack is comparable with KNN attack~\cite{knnattack} and $GeoA^3$ attack~\cite{geoa3} in attack time and much faster in total time due to the hundreds seconds of point-to-mesh reconstruction process~\cite{erler2020points2surf}.

We visualize some representative results of our Mesh Attack compared to KNN attack~\cite{knnattack} and $GeoA^3$ attack~\cite{geoa3} in Fig~\ref{final_compare}. We
can observe that for "vase" instance in ModelNet40 test set, our Mesh Attack outputs
adversarial mesh directly, while KNN attack~\cite{knnattack} and $GeoA^3$ attack~\cite{geoa3} need to reconstructed mesh from the output adversarial point cloud. Moreover, although the perturbation was small in the output adversarial point cloud, lead to very substantial "noise" in the reconstructed mesh. Those "noise" are barely 3D printable. In addition, the final adversarial point cloud sampled from the reconstructed mesh is unable to attack the victim model.

\subsection{Real physical scenario attack}
\begin{figure}
    \centering
    \includegraphics[width=\linewidth]{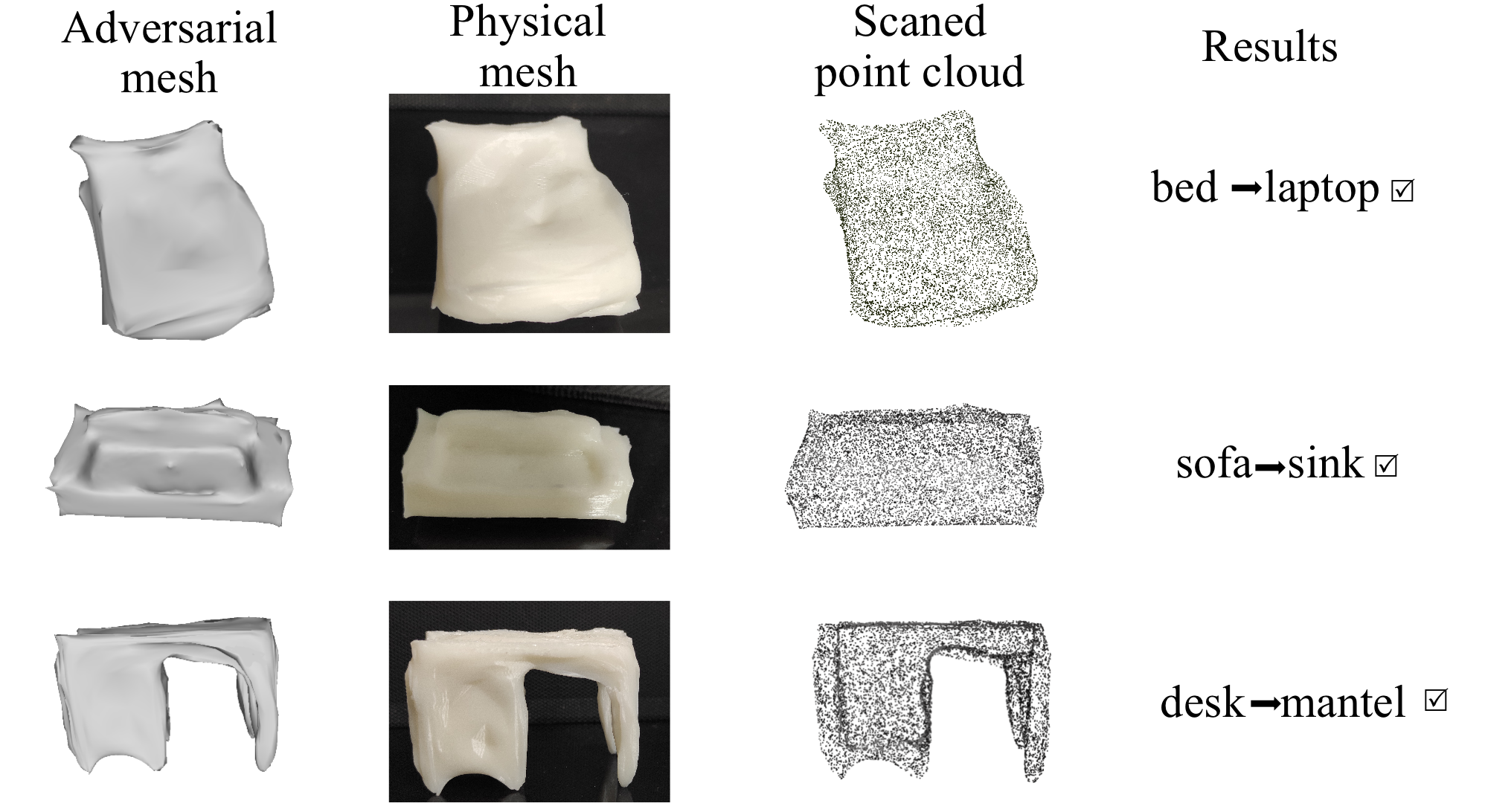}
    \caption{\textbf{Visualization of the physical attack.} Following~\cite{geoa3}, the adversarial meshes are randomly selected from Table~\ref{object_attack}. After 3D printing and scanning, the scanned points was used to attack Pointnet. A black tick indicates a successful attack.}
    \label{physical}
\end{figure}

Following~\cite{geoa3}, we randomly selected 15 instances that are achieved successful attack in Table~\ref{object_attack}. The selected meshes are printed by UP BOX Plus 3D printer and scaned by HandySCAN 700. Among them, 12 of 15 adversarial meshes can attack successfully after scanning, indicating that it is difficult to achieve a 100\% success rate in the physical scene. The Adversarial meshes, 3D printed physical meshes, and the scanned points are shown in Figure~\ref{physical}.

\begin{table*}[!t]
\centering
\setlength{\tabcolsep}{3pt}
\renewcommand\arraystretch{1.15}
\caption{Classification accuracy of various defense methods on ModelNet40 under point perturbation attack~\cite{generatingadpoint}, point dropping attack~\cite{pointcloudsaliencymaps}, KNN attack~\cite{knnattack}, AdvPC attack~\cite{advpc} and Mesh Attack on Pointnet++~\cite{pointnet++}, Pointnet~\cite{pointnet}, DGCNN~\cite{dgcnn} and PointConv~\cite{pointconv}. Drop 200 and Drop 100 denote the dropping points is 200 and 100 respectively. We report the best result of three IF-Defense.}
\label{compare_all_with_alldefense}
\vspace{2mm}
\resizebox{0.7\linewidth}{!}{
\begin{tabular}{c c c c c c c c c}
\toprule
Victim models & Defenses & Perturbation &KNN& Drop 200 & Drop 100 & AdvPC& Mesh Attack\\
\midrule
\multirow{5}*{Pointnet++} &No Defense   & \bf0.00 & \bf0.00 & 68.96 & 80.19 & 0.56&3.24$\pm0.03$ \\

&SRS  & 73.14 & 49.96 &  39.63&  64.51&48.37 &  \bf10.06$\pm0.15$\\

 &SOR & 77.67 & 61.35 & 69.17 &  74.16& 66.26& \bf5.97$\pm0.07$\\

 &DUP-Net & 80.63 & 74.88 & 72.00 & 76.38 &64.76&  \bf16.69$\pm0.33$
\\
&IF-Defense  & 86.99 & 85.62 & 79.09 & 84.56 &77.06 &\bf66.17$\pm0.26$ \\

\midrule
\multirow{5}*{Pointnet} &No Defense   & \bf0.00 & 8.51 & 40.24 & 64.67& \bf0.00 & 4.27$\pm0.01$ \\

&SRS  & 77.47 & 57.41 &  39.51&  63.57& 49.01&  \bf6.54$\pm0.05$\\

& SOR & 82.81 & 76.63 & 42.59 &  64.75& 75.45&  \bf5.73$\pm0.01$\\

& DUP-Net & 84.56 & 80.31 & 46.92 & 67.30& 77.55 &  \bf18.90$\pm0.22$
\\
&IF-Defense  & 86.30 & 86.95 & 66.94 & 77.76& 80.72 & \bf65.57$\pm0.12$ \\
\midrule
\multirow{5}*{DGCNN} &No Defense & \bf0.00 & 20.02 & 55.06 & 75.16& 9.23 & 29.21$\pm0.03$ \\
&SRS  & 50.20 & 41.25 &  \bf23.82&  49.23& 41.62&  37.03$\pm0.43$\\
 &SOR & 76.50 & 55.92 & 59.36 &  64.68& 56.49&  \bf32.10$\pm0.47$\\
 &DUP-Net & 42.67 & 35.45 & 36.02 & 44.45& \bf29.38 &  32.02$\pm0.29$
\\

&IF-Defense  & 85.53 & 82.33 & 73.30 & 84.43& 79.14 & \bf63.83$\pm0.45$ \\
\midrule
\multirow{5}*{PointConv} &No Defense   & \bf0.00 & 3.12 & 64.02 & 77.96& 6.45 & 8.02$\pm0.03$ \\
&SRS  & 76.22 & 55.75 &  48.87&  69.45& 37.62&  \bf10.93$\pm0.25$\\

 &SOR & 79.25 & 26.13 & 63.78 &  77.63& 51.75&  \bf10.13$\pm0.24$\\

 &DUP-Net & 68.84 & 43.76 & 58.23 & 70.75 & 49.35&  \bf13.39$\pm0.04$
\\
&IF-Defense  & 86.67 & 81.08 & 74.51 & 81.20& 61.77 & \bf53.10$\pm0.78$ \\
\bottomrule
\end{tabular}
}
\end{table*}

\subsection{Mesh Attack under Defense}
We further verified our Mesh Attack under various defenses. Following the experiment detail in \cite{ifdefense}, we performed a targeted attack of Mesh Attack, the target label of each instance is the same as \cite{ifdefense} for a fair comparison. We compared our Mesh Attack with the point perturbation attack~\cite{generatingadpoint}, KNN attack~\cite{knnattack}, Drop point attack~\cite{pointcloudsaliencymaps} and AdvPC attack~\cite{advpc} under various defense algorithms. Since our Mesh Attack employed in the whole test set, the point cloud that the victim model fails or success to classify are counted in the statistics of our attack performance, therefore, we use the classification accuracy as the evaluation metrics. The defense method of Simple Random Sampling (SRS), Statistical Outlier Removal (SOR), DUP-Net defense~\cite{dupnet} and IF-Defense~\cite{ifdefense} are used for evaluating the adversarial robustness of various attacks. 
The results of the point perturbation attack~\cite{generatingadpoint}, KNN attack~\cite{knnattack}, Drop point attack~\cite{pointcloudsaliencymaps} and AdvPC attack~\cite{advpc} under various defenses are refer to IF-Defense~\cite{ifdefense}.
As shown in Table \ref{compare_all_with_alldefense}, our Mesh Attack consistently outperform other attacks under various defense algorithms. The SRS, SOR and DUP-Net have almost no defense effect on our Mesh Attack. This is because our adversarial point cloud examples are directly sampled from the surface of the mesh, and there are almost no outliers.
Our Mesh Attack achieved state-of-the-art performance across various defense algorithms, therefore, Mesh Attack is the hardest to defend. More qualitative comparisons of Mesh Attack, perturbation attack~\cite{generatingadpoint}, KNN attack~\cite{knnattack} and $GeoA^3$ attack~\cite{geoa3} are refer to the supplementary material.

\subsection{Black-box attack on the Mesh Classifier}

The adversarial mesh examples generated in Sec. 4.5. are used to perform a black-box attack on the MeshNet. The victim model is the pre-trained model of MeshNet with 91.92\% accuracy in the ModelNet40 test set, which was kindly provided by MeshNet~\cite{feng2019meshnet}. The results are shown in Table \ref{transmesh}. We report the classification accuracy of MeshNet to classify the adversarial mesh examples generated by Pointnet, Pointnet++, DGCNN, and PointConv. The results verified that our Mesh Attack could be generalized to the mesh classifier. To the best of our knowledge, it was the first adversarial attack to mesh classifier, which showed that adversarial examples can fool the mesh classifier and show some threat to DNN based mesh research in the future.

\begin{table}[!t]
\begin{center}
\centering
\caption{The classification accuracy (\%) of untargeted Mesh Attack transfer to MeshNet~\cite{feng2019meshnet}}
\label{transmesh}
\footnotesize
\resizebox{\linewidth}{!}{
\begin{tabular}{ccccc}
\toprule
Model & Pointnet & Pointnet++ & DGCNN & PointConv \\
\midrule
 Clean&\multicolumn{4}{c}{91.92}  \\
 \midrule
 Mesh Attack&27.43$\pm0.14$ &27.19$\pm0.19$ & \bf25.49$\pm0.08$ & 30.92$\pm0.11$   \\
\bottomrule
\end{tabular}
}
\end{center}
\vspace{-1ex}
\end{table}



\section{Conclusions}

In this paper, we proposed a strong adversarial attack for point cloud and mesh, named Mesh Attack, which updated the mesh vertices in each iteration of the attack, and utilized three mesh losses to regularize the mesh to be smooth.
The experiment results showed that our Mesh Attack outperforms the existing point cloud attack methods by a large margin and hard to be defended by the current state-of-the-art point cloud defense algorithms.
We also perform a black-box attack on the mesh classifier, and the results are promising.
This raises a new question for the current adversarial robustness research: Do we really need to perform adversarial attacks in the input space?
With the huge breakthrough made by NeRF~\cite{mildenhall2020nerf} in scene representation, we believed that an attack in this kind of scene representation would be promising.
Moreover, effective defense methods against Mesh Attack will be another crucial and promising direction. In addition, unlike image domain, there is a lack of the perceptual metric which measures how similar the two objects are in a way that we humans do in 3D area, neither point cloud nor mesh, which will limit the research on 3D adversarial attack and defense.

{
    \small
    \bibliographystyle{ieee_fullname}
    \bibliography{macros,main}

\begin{thebibliography}{10}\itemsep=-1pt

\bibitem{adsurvey}
Naveed Akhtar and Ajmal Mian.
\newblock Threat of adversarial attacks on deep learning in computer vision: A
  survey.
\newblock {\em IEEE Access}, 6:14410--14430, 2018.

\bibitem{athalye2018synthesizingphy}
Anish Athalye, Logan Engstrom, Andrew Ilyas, and Kevin Kwok.
\newblock Synthesizing robust adversarial examples.
\newblock In {\em International conference on machine learning}, pages
  284--293. PMLR, 2018.

\bibitem{brown2017adversarial_patch_phy}
Tom~B Brown, Dandelion Man{\'e}, Aurko Roy, Mart{\'\i}n Abadi, and Justin
  Gilmer.
\newblock Adversarial patch.
\newblock {\em arXiv preprint arXiv:1712.09665}, 2017.

\bibitem{cao2019adversarialcars3d}
Yulong Cao, Chaowei Xiao, Benjamin Cyr, Yimeng Zhou, Won Park, Sara Rampazzi,
  Qi~Alfred Chen, Kevin Fu, and Z~Morley Mao.
\newblock Adversarial sensor attack on lidar-based perception in autonomous
  driving.
\newblock In {\em Proceedings of the 2019 ACM SIGSAC conference on computer and
  communications security}, pages 2267--2281, 2019.

\bibitem{cwattack}
Nicholas Carlini and David Wagner.
\newblock Towards evaluating the robustness of neural networks.
\newblock In {\em 2017 ieee symposium on security and privacy (sp)}, pages
  39--57. IEEE, 2017.

\bibitem{zooattackblack}
Pin-Yu Chen, Huan Zhang, Yash Sharma, Jinfeng Yi, and Cho-Jui Hsieh.
\newblock Zoo: Zeroth order optimization based black-box attacks to deep neural
  networks without training substitute models.
\newblock In {\em Proceedings of the 10th ACM workshop on artificial
  intelligence and security}, pages 15--26, 2017.

\bibitem{mifgm}
Yinpeng Dong, Fangzhou Liao, Tianyu Pang, Hang Su, Jun Zhu, Xiaolin Hu, and
  Jianguo Li.
\newblock Boosting adversarial attacks with momentum.
\newblock In {\em Proceedings of the IEEE conference on computer vision and
  pattern recognition}, pages 9185--9193, 2018.

\bibitem{camouflagephy}
Ranjie Duan, Xingjun Ma, Yisen Wang, James Bailey, A~Kai Qin, and Yun Yang.
\newblock Adversarial camouflage: Hiding physical-world attacks with natural
  styles.
\newblock In {\em Proceedings of the IEEE/CVF Conference on Computer Vision and
  Pattern Recognition}, pages 1000--1008, 2020.

\bibitem{erler2020points2surf}
Philipp Erler, Paul Guerrero, Stefan Ohrhallinger, Niloy~J Mitra, and Michael
  Wimmer.
\newblock Points2surf learning implicit surfaces from point clouds.
\newblock In {\em European Conference on Computer Vision}, pages 108--124.
  Springer, 2020.

\bibitem{robustphy}
Kevin Eykholt, Ivan Evtimov, Earlence Fernandes, Bo Li, Amir Rahmati, Chaowei
  Xiao, Atul Prakash, Tadayoshi Kohno, and Dawn Song.
\newblock Robust physical-world attacks on deep learning visual classification.
\newblock In {\em Proceedings of the IEEE Conference on Computer Vision and
  Pattern Recognition}, pages 1625--1634, 2018.

\bibitem{feng2019meshnet}
Yutong Feng, Yifan Feng, Haoxuan You, Xibin Zhao, and Yue Gao.
\newblock Meshnet: Mesh neural network for 3d shape representation.
\newblock In {\em Proceedings of the AAAI Conference on Artificial
  Intelligence}, volume~33, pages 8279--8286, 2019.

\bibitem{goodfellowgan}
Ian~J Goodfellow, Jean Pouget-Abadie, Mehdi Mirza, Bing Xu, David Warde-Farley,
  Sherjil Ozair, Aaron Courville, and Yoshua Bengio.
\newblock Generative adversarial networks.
\newblock {\em arXiv preprint arXiv:1406.2661}, 2014.

\bibitem{fgm}
Ian~J Goodfellow, Jonathon Shlens, and Christian Szegedy.
\newblock Explaining and harnessing adversarial examples.
\newblock {\em arXiv preprint arXiv:1412.6572}, 2014.

\bibitem{advpc}
Abdullah Hamdi, Sara Rojas, Ali Thabet, and Bernard Ghanem.
\newblock Advpc: Transferable adversarial perturbations on 3d point clouds.
\newblock In {\em European Conference on Computer Vision}, pages 241--257.
  Springer, 2020.

\bibitem{manifold40}
Shi-Min Hu, Zheng-Ning Liu, Meng-Hao Guo, Jun-Xiong Cai, Jiahui Huang,
  Tai-Jiang Mu, and Ralph~R Martin.
\newblock Subdivision-based mesh convolution networks.
\newblock {\em arXiv preprint arXiv:2106.02285}, 2021.

\bibitem{ifgm}
Alexey Kurakin, Ian Goodfellow, and Samy Bengio.
\newblock Adversarial machine learning at scale.
\newblock {\em arXiv preprint arXiv:1611.01236}, 2016.

\bibitem{adv_inphy}
Alexey Kurakin, Ian Goodfellow, Samy Bengio, et~al.
\newblock Adversarial examples in the physical world, 2016.

\bibitem{lambourne2021brepnet}
Joseph~G. Lambourne, Karl~D.D. Willis, Pradeep~Kumar Jayaraman, Aditya Sanghi,
  Peter Meltzer, and Hooman Shayani.
\newblock Brepnet: A topological message passing system for solid models.
\newblock In {\em IEEE Conference on Computer Vision and Pattern Recognition
  (CVPR)}, 2021.

\bibitem{rscnn}
Yongcheng Liu, Bin Fan, Shiming Xiang, and Chunhong Pan.
\newblock Relation-shape convolutional neural network for point cloud analysis.
\newblock In {\em Proceedings of the IEEE Conference on Computer Vision and
  Pattern Recognition}, pages 8895--8904, 2019.

\bibitem{attsordefense}
Chengcheng Ma, Weiliang Meng, Baoyuan Wu, Shibiao Xu, and Xiaopeng Zhang.
\newblock Efficient joint gradient based attack against sor defense for 3d
  point cloud classification.
\newblock In {\em Proceedings of the 28th ACM International Conference on
  Multimedia}, pages 1819--1827, 2020.

\bibitem{pgd}
Aleksander Madry, Aleksandar Makelov, Ludwig Schmidt, Dimitris Tsipras, and
  Adrian Vladu.
\newblock Towards deep learning models resistant to adversarial attacks.
\newblock {\em arXiv preprint arXiv:1706.06083}, 2017.

\bibitem{mildenhall2020nerf}
Ben Mildenhall, Pratul~P Srinivasan, Matthew Tancik, Jonathan~T Barron, Ravi
  Ramamoorthi, and Ren Ng.
\newblock Nerf: Representing scenes as neural radiance fields for view
  synthesis.
\newblock In {\em European Conference on Computer Vision}, pages 405--421.
  Springer, 2020.

\bibitem{moosavi2017universal}
Seyed-Mohsen Moosavi-Dezfooli, Alhussein Fawzi, Omar Fawzi, and Pascal
  Frossard.
\newblock Universal adversarial perturbations.
\newblock In {\em Proceedings of the IEEE conference on computer vision and
  pattern recognition}, pages 1765--1773, 2017.

\bibitem{moosavi2016deepfool}
Seyed-Mohsen Moosavi-Dezfooli, Alhussein Fawzi, and Pascal Frossard.
\newblock Deepfool: a simple and accurate method to fool deep neural networks.
\newblock In {\em Proceedings of the IEEE conference on computer vision and
  pattern recognition}, pages 2574--2582, 2016.

\bibitem{paszke2019pytorch}
Adam Paszke, Sam Gross, Francisco Massa, Adam Lerer, James Bradbury, Gregory
  Chanan, Trevor Killeen, Zeming Lin, Natalia Gimelshein, Luca Antiga, et~al.
\newblock Pytorch: An imperative style, high-performance deep learning library.
\newblock {\em Advances in neural information processing systems},
  32:8026--8037, 2019.

\bibitem{pointnet}
Charles~R Qi, Hao Su, Kaichun Mo, and Leonidas~J Guibas.
\newblock Pointnet: Deep learning on point sets for 3d classification and
  segmentation.
\newblock In {\em Proceedings of the IEEE conference on computer vision and
  pattern recognition}, pages 652--660, 2017.

\bibitem{qi2016volumetric_multi}
Charles~R Qi, Hao Su, Matthias Nie{\ss}ner, Angela Dai, Mengyuan Yan, and
  Leonidas~J Guibas.
\newblock Volumetric and multi-view cnns for object classification on 3d data.
\newblock In {\em Proceedings of the IEEE conference on computer vision and
  pattern recognition}, pages 5648--5656, 2016.

\bibitem{pointnet++}
Charles~Ruizhongtai Qi, Li Yi, Hao Su, and Leonidas~J Guibas.
\newblock Pointnet++: Deep hierarchical feature learning on point sets in a
  metric space.
\newblock {\em Advances in neural information processing systems},
  30:5099--5108, 2017.

\bibitem{adreview}
Shilin Qiu, Qihe Liu, Shijie Zhou, and Chunjiang Wu.
\newblock Review of artificial intelligence adversarial attack and defense
  technologies.
\newblock {\em Applied Sciences}, 9(5):909, 2019.

\bibitem{ravi2020pytorch3d}
Nikhila Ravi, Jeremy Reizenstein, David Novotny, Taylor Gordon, Wan-Yen Lo,
  Justin Johnson, and Georgia Gkioxari.
\newblock Accelerating 3d deep learning with pytorch3d.
\newblock {\em arXiv:2007.08501}, 2020.

\bibitem{sharif2016accessorizephy}
Mahmood Sharif, Sruti Bhagavatula, Lujo Bauer, and Michael~K Reiter.
\newblock Accessorize to a crime: Real and stealthy attacks on state-of-the-art
  face recognition.
\newblock In {\em Proceedings of the 2016 acm sigsac conference on computer and
  communications security}, pages 1528--1540, 2016.

\bibitem{pvrcnn}
Shaoshuai Shi, Chaoxu Guo, Li Jiang, Zhe Wang, Jianping Shi, Xiaogang Wang, and
  Hongsheng Li.
\newblock Pv-rcnn: Point-voxel feature set abstraction for 3d object detection.
\newblock In {\em Proceedings of the IEEE/CVF Conference on Computer Vision and
  Pattern Recognition}, pages 10529--10538, 2020.

\bibitem{su2015multi}
Hang Su, Subhransu Maji, Evangelos Kalogerakis, and Erik Learned-Miller.
\newblock Multi-view convolutional neural networks for 3d shape recognition.
\newblock In {\em Proceedings of the IEEE international conference on computer
  vision}, pages 945--953, 2015.

\bibitem{su2019one}
Jiawei Su, Danilo~Vasconcellos Vargas, and Kouichi Sakurai.
\newblock One pixel attack for fooling deep neural networks.
\newblock {\em IEEE Transactions on Evolutionary Computation}, 23(5):828--841,
  2019.

\bibitem{firstadv}
Christian Szegedy, Wojciech Zaremba, Ilya Sutskever, Joan Bruna, Dumitru Erhan,
  Ian Goodfellow, and Rob Fergus.
\newblock Intriguing properties of neural networks.
\newblock {\em arXiv preprint arXiv:1312.6199}, 2013.

\bibitem{knnattack}
Tzungyu Tsai, Kaichen Yang, Tsung-Yi Ho, and Yier Jin.
\newblock Robust adversarial objects against deep learning models.
\newblock In {\em Proceedings of the AAAI Conference on Artificial
  Intelligence}, volume~34, pages 954--962, 2020.

\bibitem{tu2019autozoomblack}
Chun-Chen Tu, Paishun Ting, Pin-Yu Chen, Sijia Liu, Huan Zhang, Jinfeng Yi,
  Cho-Jui Hsieh, and Shin-Ming Cheng.
\newblock Autozoom: Autoencoder-based zeroth order optimization method for
  attacking black-box neural networks.
\newblock In {\em Proceedings of the AAAI Conference on Artificial
  Intelligence}, volume~33, pages 742--749, 2019.

\bibitem{tu2020physically3d}
James Tu, Mengye Ren, Sivabalan Manivasagam, Ming Liang, Bin Yang, Richard Du,
  Frank Cheng, and Raquel Urtasun.
\newblock Physically realizable adversarial examples for lidar object
  detection.
\newblock In {\em Proceedings of the IEEE/CVF Conference on Computer Vision and
  Pattern Recognition}, pages 13716--13725, 2020.

\bibitem{wang2018pixel2mesh}
Nanyang Wang, Yinda Zhang, Zhuwen Li, Yanwei Fu, Wei Liu, and Yu-Gang Jiang.
\newblock Pixel2mesh: Generating 3d mesh models from single rgb images.
\newblock In {\em Proceedings of the European Conference on Computer Vision
  (ECCV)}, pages 52--67, 2018.

\bibitem{wang2020pixel2mesh}
Nanyang Wang, Yinda Zhang, Zhuwen Li, Yanwei Fu, Hang Yu, Wei Liu, Xiangyang
  Xue, and Yu-Gang Jiang.
\newblock Pixel2mesh: 3d mesh model generation via image guided deformation.
\newblock {\em IEEE transactions on pattern analysis and machine intelligence},
  2020.

\bibitem{dgcnn}
Yue Wang, Yongbin Sun, Ziwei Liu, Sanjay~E Sarma, Michael~M Bronstein, and
  Justin~M Solomon.
\newblock Dynamic graph cnn for learning on point clouds.
\newblock {\em Acm Transactions On Graphics (tog)}, 38(5):1--12, 2019.

\bibitem{wen2019pixel2mesh++}
Chao Wen, Yinda Zhang, Zhuwen Li, and Yanwei Fu.
\newblock Pixel2mesh++: Multi-view 3d mesh generation via deformation.
\newblock In {\em Proceedings of the IEEE/CVF International Conference on
  Computer Vision}, pages 1042--1051, 2019.

\bibitem{geoa3}
Yuxin Wen, Jiehong Lin, Ke Chen, CL~Philip Chen, and Kui Jia.
\newblock Geometry-aware generation of adversarial point clouds.
\newblock {\em IEEE Transactions on Pattern Analysis and Machine Intelligence},
  2020.

\bibitem{pointconv}
Wenxuan Wu, Zhongang Qi, and Li Fuxin.
\newblock Pointconv: Deep convolutional networks on 3d point clouds.
\newblock In {\em Proceedings of the IEEE/CVF Conference on Computer Vision and
  Pattern Recognition}, pages 9621--9630, 2019.

\bibitem{ifdefense}
Ziyi Wu, Yueqi Duan, He Wang, Qingnan Fan, and Leonidas~J Guibas.
\newblock If-defense: 3d adversarial point cloud defense via implicit function
  based restoration.
\newblock {\em arXiv preprint arXiv:2010.05272}, 2020.

\bibitem{modelnet40}
Zhirong Wu, Shuran Song, Aditya Khosla, Fisher Yu, Linguang Zhang, Xiaoou Tang,
  and Jianxiong Xiao.
\newblock 3d shapenets: A deep representation for volumetric shapes.
\newblock In {\em Proceedings of the IEEE conference on computer vision and
  pattern recognition}, pages 1912--1920, 2015.

\bibitem{generatingadpoint}
Chong Xiang, Charles~R Qi, and Bo Li.
\newblock Generating 3d adversarial point clouds.
\newblock In {\em Proceedings of the IEEE Conference on Computer Vision and
  Pattern Recognition}, pages 9136--9144, 2019.

\bibitem{tshirt_phy}
Kaidi Xu, Gaoyuan Zhang, Sijia Liu, Quanfu Fan, Mengshu Sun, Hongge Chen,
  Pin-Yu Chen, Yanzhi Wang, and Xue Lin.
\newblock Adversarial t-shirt! evading person detectors in a physical world.
\newblock In {\em European Conference on Computer Vision}, pages 665--681.
  Springer, 2020.

\bibitem{pointasnl}
Xu Yan, Chaoda Zheng, Zhen Li, Sheng Wang, and Shuguang Cui.
\newblock Pointasnl: Robust point clouds processing using nonlocal neural
  networks with adaptive sampling.
\newblock In {\em Proceedings of the IEEE/CVF Conference on Computer Vision and
  Pattern Recognition}, pages 5589--5598, 2020.

\bibitem{yuan2019adversarial}
Xiaoyong Yuan, Pan He, Qile Zhu, and Xiaolin Li.
\newblock Adversarial examples: Attacks and defenses for deep learning.
\newblock {\em IEEE transactions on neural networks and learning systems},
  30(9):2805--2824, 2019.

\bibitem{pointcutmix}
Jinlai Zhang, Lvjie Chen, Bo Ouyang, Binbin Liu, Jihong Zhu, Yujing Chen,
  Yanmei Meng, and Danfeng Wu.
\newblock Pointcutmix: Regularization strategy for point cloud classification.
\newblock {\em arXiv preprint arXiv:2101.01461}, 2021.

\bibitem{advnlp}
Wei~Emma Zhang, Quan~Z Sheng, Ahoud Alhazmi, and Chenliang Li.
\newblock Adversarial attacks on deep-learning models in natural language
  processing: A survey.
\newblock {\em ACM Transactions on Intelligent Systems and Technology (TIST)},
  11(3):1--41, 2020.

\bibitem{pointcloudsaliencymaps}
Tianhang Zheng, Changyou Chen, Junsong Yuan, Bo Li, and Kui Ren.
\newblock Pointcloud saliency maps.
\newblock In {\em Proceedings of the IEEE International Conference on Computer
  Vision}, pages 1598--1606, 2019.

\bibitem{lggan}
Hang Zhou, Dongdong Chen, Jing Liao, Kejiang Chen, Xiaoyi Dong, Kunlin Liu,
  Weiming Zhang, Gang Hua, and Nenghai Yu.
\newblock Lg-gan: Label guided adversarial network for flexible targeted attack
  of point cloud based deep networks.
\newblock In {\em Proceedings of the IEEE/CVF Conference on Computer Vision and
  Pattern Recognition}, pages 10356--10365, 2020.

\bibitem{dupnet}
Hang Zhou, Kejiang Chen, Weiming Zhang, Han Fang, Wenbo Zhou, and Nenghai Yu.
\newblock Dup-net: Denoiser and upsampler network for 3d adversarial point
  clouds defense.
\newblock In {\em Proceedings of the IEEE International Conference on Computer
  Vision}, pages 1961--1970, 2019.

\end{thebibliography}
}

\appendix

\setcounter{page}{1}

\twocolumn[
\centering
\Large
\vspace{0.5em}Supplementary Material \\
\vspace{1.0em}
] 
\appendix

\section{Qualitative results}
\begin{figure*}
    \centering
    \includegraphics[width=\textwidth]{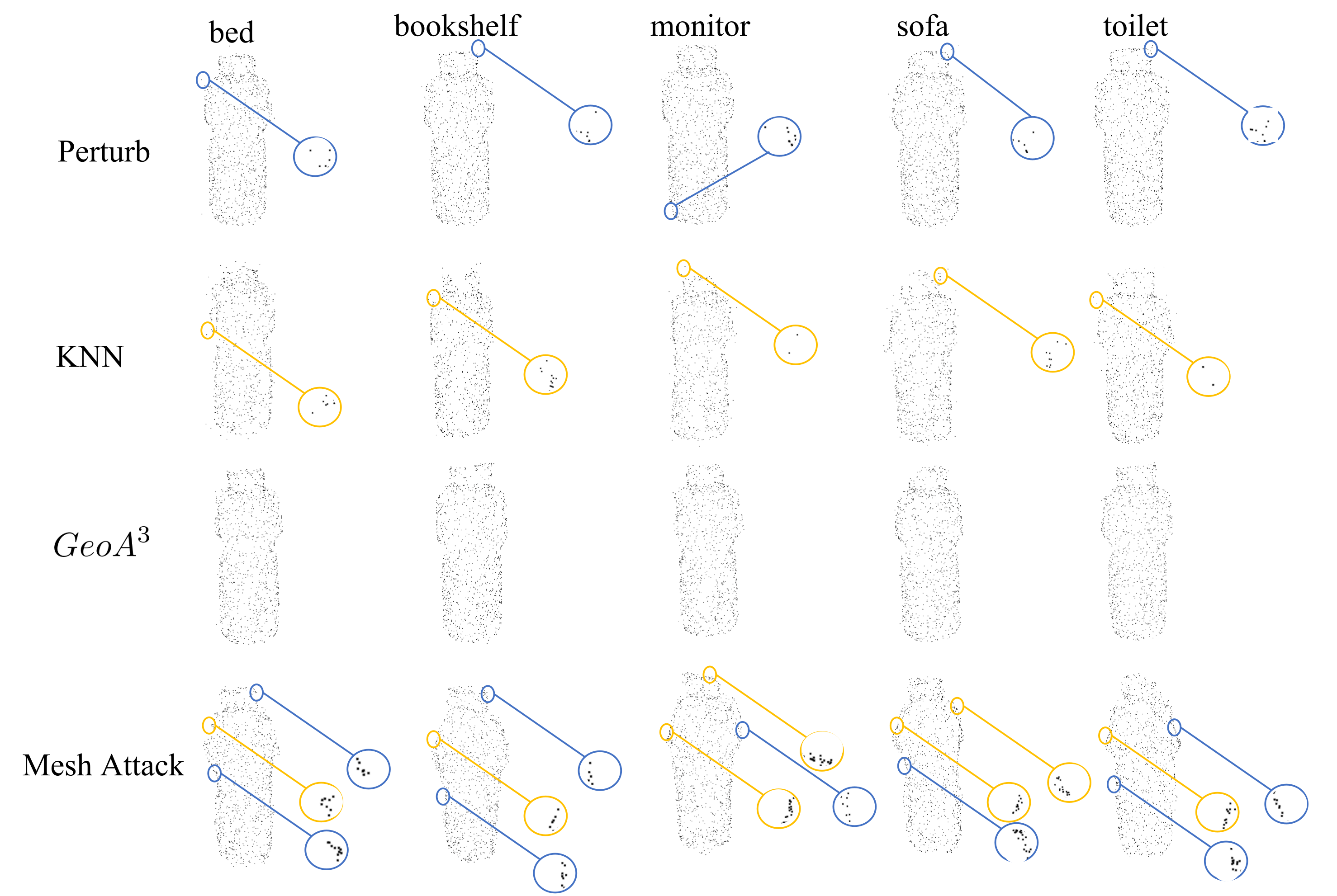}
    \caption{Qualitative comparisons of different adversarial point cloud attack algorithms. From top row to the bottom row are point perturbation attack~\cite{generatingadpoint}, KNN attack~\cite{knnattack}, $GeoA^3$ attack~\cite{geoa3} and Mesh Attack. We perform the targetted attack with bottle instance to Pointnet~\cite{pointnet}, the targets are bed, bookshelf, monitor, sofa and toilet from left to right.}
    \label{pl_compare}
\end{figure*}

We visualize some adversarial point cloud examples to analyze the adversarial robustness of Mesh Attack. As shown in Figure~\ref{pl_compare}, the adversarial effects of perturbation attack~\cite{generatingadpoint} and KNN attack~\cite{knnattack} mainly come from outliers. Although $GeoA^3$ attack~\cite{geoa3} has fewer outliers, the adversarial effect drops sharply after meshing and re-sampling, indicating that its adversarial effect mainly depends on specific point cloud space. It can be seen from the figure that the adversarial effect of Mesh Attack mainly comes from the deformation of the object, and because our adversarial point cloud is directly sampled on the smooth adversarial mesh surface, our adversarial point cloud has almost no outliers.

\end{document}